\documentclass[10pt,twocolumn,letterpaper]{article}

\usepackage{cvpr}
\usepackage{amsmath}
\usepackage{amsfonts}
\usepackage{amssymb}
\usepackage{times}
\usepackage{graphicx}
\usepackage{caption}
\usepackage{subcaption}
\usepackage{textcomp}
\usepackage[activate]{pdfcprot}
\usepackage{fixltx2e}
\usepackage{tabularx}
\newcolumntype{P}[1]{>{\raggedright\arraybackslash}m{#1}}%
\newcolumntype{C}[1]{>{\centering\arraybackslash}m{#1}}%
\newcolumntype{R}[1]{>{\raggedleft\arraybackslash}m{#1}}%

\usepackage[pagebackref=true,breaklinks=true,colorlinks,bookmarks=false]{hyperref}
\newcommand{\owntitle}{Prediction of Search Targets From Fixations in Open-World Settings}
\newcommand{\ownkeywords}{Search target prediction, eye tracking, mental image}

\cvprfinalcopy

\ifcvprfinal\fi

\DeclareMathOperator*{\argmax}{arg\,max}
\hypersetup{%
pdftitle={\owntitle},
pdfauthor={Hosnieh Sattar, Sabine M\"uller, Mario Fritz, Andreas Bulling},
pdfkeywords={\ownkeywords},
bookmarksnumbered,
pdfstartview={FitH},
colorlinks,
citecolor=red,
filecolor=red,
linkcolor=red,
urlcolor=red,
breaklinks=true
}
\begin{document}

\title{\owntitle}

\author{
Hosnieh Sattar$^{1,2}$
\and
Sabine M\"uller$^1$
\and
Mario Fritz$^2$
\and
Andreas Bulling$^1$\\
\and
$^1$Perceptual User Interfaces Group, $^2$Scalable Learning and Perception Group\\
Max Planck Institute for Informatics, Saarbr\"ucken, Germany\\
{\tt\small \{sattar,smueller,mfritz,bulling\}@mpi-inf.mpg.de}
}

\maketitle
\begin{abstract}
Previous work on predicting the target of visual search from human fixations only considered closed-world settings in which training labels are available and predictions are performed for a known set of potential targets.
In this work we go beyond the state of the art by studying search target prediction in an open-world setting in which we no longer assume that we have fixation data to train for the search targets.
We present a dataset containing fixation data of 18 users searching for natural images from three image categories within synthesised image collages of about 80 images.
In a closed-world baseline experiment we show that we can predict the correct target image out of a candidate set of five images. We then present a new problem formulation for search target prediction in the open-world setting that is based on learning compatibilities between fixations and potential targets.
\end{abstract}

\section{Introduction}

\begin{figure}[t]
\centering
\includegraphics[width=1\columnwidth]{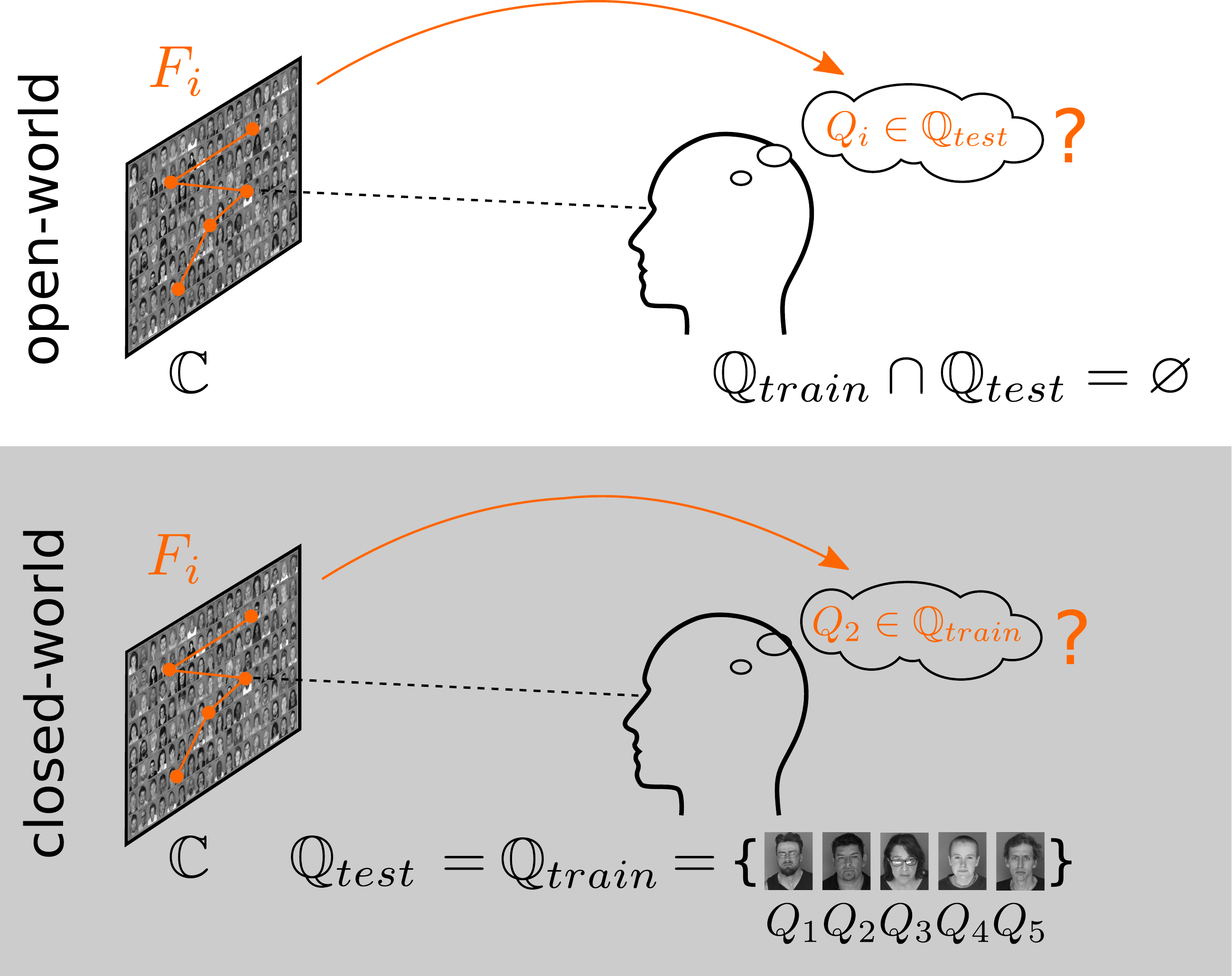}
\caption{Experiments conducted in this work. In the \textit{closed-world} experiment we aim to predict which target image (here $Q_2$) out of a candidate set of five images $\mathbb{Q}_{train} = \mathbb{Q}_{test}$ the user is searching for by analysing fixations $F_i$ on an image collage $\mathbb{C}$. In the \textit{open-world} experiments we aim to predict $Q_i$ on the whole $\mathbb{Q}_{test}$.}
\label{fig:evaluations}
\end{figure}

In his seminal work from 1967, Yarbus showed that visual behaviour is closely linked to task when looking at a visual scene~\cite{yarbus1967eye}. This work is an important demonstration of task influence on fixation patterns and sparked a large number of follow-up works in a range of disciplines, 
including human vision, neuroscience, artificial intelligence, and computer vision. A common goal in these human and computer vision works is to analyse visual behaviour, i.e.\ typically 
fixations and saccades, in order to make predictions about user behaviour. For example, previous work has used visual behaviour analysis as a means to predict the 
users' tasks~\cite{borji2014defending, borji2012probabilistic, deangelus2009top, haji2014inverse,kanan:predicting,zelinsky2013eye,ZhangNIPS2005b}, 
visual activities~\cite{bulling12_tap,bulling11_pami,bulling13_chi, Land2006296,Peters_Itti08nips}, 
cognitive processes such as memory recall or high cognitive load~\cite{bulling11_ubicomp,tessendorf11_pervasive}, abstract thought processes~\cite{coen2009visuomotor,mast2002eye}, 
the type of a visual stimulus~\cite{brandt1997spontaneous,moran,994097}, interest for interactive image retrieval~\cite{6152440,guo2002learning,Hussain,Kozma09, papadopoulos2013gaze,stefanoumental,Zhang}, which number a person has in mind~\cite{loetscher2010eye}, or -- most recently -- to predict the search target during visual search~\cite{borji2014eyes, haji2013computational, rajashekar2006visual,zelinsky2013eye}.

Predicting the target of a visual search task is particularly interesting, as the corresponding internal representation, the mental image of the search target, is difficult if not impossible to assess using other modalities. While~\cite{zelinsky2013eye} and~\cite{borji2014eyes} underlined the significant potential of using gaze information to predict visual search targets, they both considered a closed-world setting. In this setting, all potential search targets are part of the training set, and fixations for all of these targets were observed.

In contrast, in this work we study an open-world setting in which we no longer assume that we have fixation data to train for these targets. Search target prediction in this setting has significant practical relevance for a range of applications, such as image and media retrieval. This setting is challenging because we have to develop a learning mechanism that can predict over an unknown set of targets. We study this problem on a new dataset that contains fixation data of 18 users searching for five target images from three categories (faces as well as two different sets of book covers) in collages synthesised from about 80 images. The dataset is publicly available online.

The contributions of this work are threefold. First, we present an annotated dataset of human fixations on synthesised collages of natural images during visual search that lends itself to studying our new open-world setting. 
Compared to previous works, our dataset is more challenging because of its larger number of distractors, higher similarities between search image and distractors, and a larger number of potential search targets.
Second, we introduce a novel problem formulation and method for learning the compatibility between observed fixations and potential search targets.
Third, using this dataset, we report on a series of experiments on predicting users' search target from fixations by moving from closed-world to open-world settings.

\section{Related Work}

Our work is related to previous works on analysing gaze information in order to make predictions about general user behaviour as well as on predicting search targets from fixations during visual search tasks.

\subsection{Predicting User Behaviour From Gaze}

Several researchers recently aimed to reproduce Yarbus's findings and to extend them by automatically predicting the observers' tasks. Green et al.\ reproduced the original experiments, but although they were able to predict the observers' identity and the observed images from the scanpaths, they did not succeed in predicting the task itself~\cite{greene2012reconsidering}. Borji et al., Kanan et al., and Haji-Abolhassani et al.\ conducted follow-up experiments using more sophisticated features and machine learning techniques~\cite{borji2014defending, haji2014inverse, kanan:predicting}. All three works showed that the observers' tasks could be successfully predicted from gaze information alone.

Other works investigated means to recognise more general aspects of user behaviour. Bulling et al.\ investigated the recognition of everyday office activities from visual behaviour, such as reading, taking hand-written notes, or browsing the web~\cite{bulling11_pami}. Based on long-term eye movement recordings, they later showed that high-level contextual cues, such as social interactions or being mentally active, could also be inferred from visual behaviour~\cite{bulling13_chi}. They further showed that cognitive processes, such as visual memory recall or cognitive load, could be inferred from gaze information~\cite{bulling11_ubicomp,tessendorf11_pervasive} as well -- of which the former finding was recently confirmed by Henderson et al.~\cite{henderson2013predicting}.

Several previous works investigated the use of gaze information as an implicit measure of relevance in image retrieval tasks. For example, Oyekoya and Stendiford compared similarity measures based on a visual saliency model as well as real human gaze patterns, indicating better performance for gaze~\cite{oyekoya2004eye}. In later works the same and other authors showed that gaze information yielded significantly better performance than random selection or using saliency information~\cite{oyekoya2007perceptual,schulze2013eye}. Coddington presented a similar system but used two separate screens for the task~\cite{6152440} while Kozma et al.\ focused on implicit cues obtained from gaze in real-time interfaces~\cite{Kozma09}. With the goal of making implicit relevance feedback richer, Klami proposed to infer which parts of the image the user found most relevant from gaze~\cite{klami2010inferring}.

\subsection{Predicting Search Targets From Gaze}

Only a few previous works here focused on visual search and the problem of predicting search targets from gaze.
Zelinsky et al.\ aimed to predict subjects' gaze patterns during categorical search tasks~\cite{zelinsky2013modelling}. 
They designed a series of experiments in which participants had to find two categorical search targets (teddy bear and butterfly) among four visually similar distractors. They predicted the number of fixations made prior to search judgements as well as the percentage of first eye movements landing on the search target. In another work they showed how to predict the categorical search targets themselves from eye fixations~\cite{zelinsky2013eye}. Borji et al.\ focused on predicting search targets from fixations~\cite{borji2014eyes}. In three experiments, participants had to find a binary pattern and 3-level luminance patterns out of a set of other patterns, 
as well as one of 15 objects in 11 synthetic natural scenes. They showed that binary patterns with higher similarity to the search target were viewed more often by participants. Additionally, they found that when the complexity of the search target increased, participants were guided more by sub-patterns rather than the whole pattern. 

\subsection{Summary}

The works of Zelinsky et al.~\cite{zelinsky2013eye} and Borji et al.~\cite{borji2014eyes} are most related to ours. However, both works only considered simplified visual stimuli or synthesised natural scenes in a closed-world setting. In that setting, all potential search targets were part of the training set and fixations for all of these targets were observed. In contrast, our work is the first to address the open-world setting in which we no longer assume that we have fixation data to train for these targets, and to present a new problem formulation for this open-world search target recognition in the open-world setting.

\section{Data Collection and Collage Synthesis}

\begin{figure}[t]
\centering
\includegraphics[width=\columnwidth]{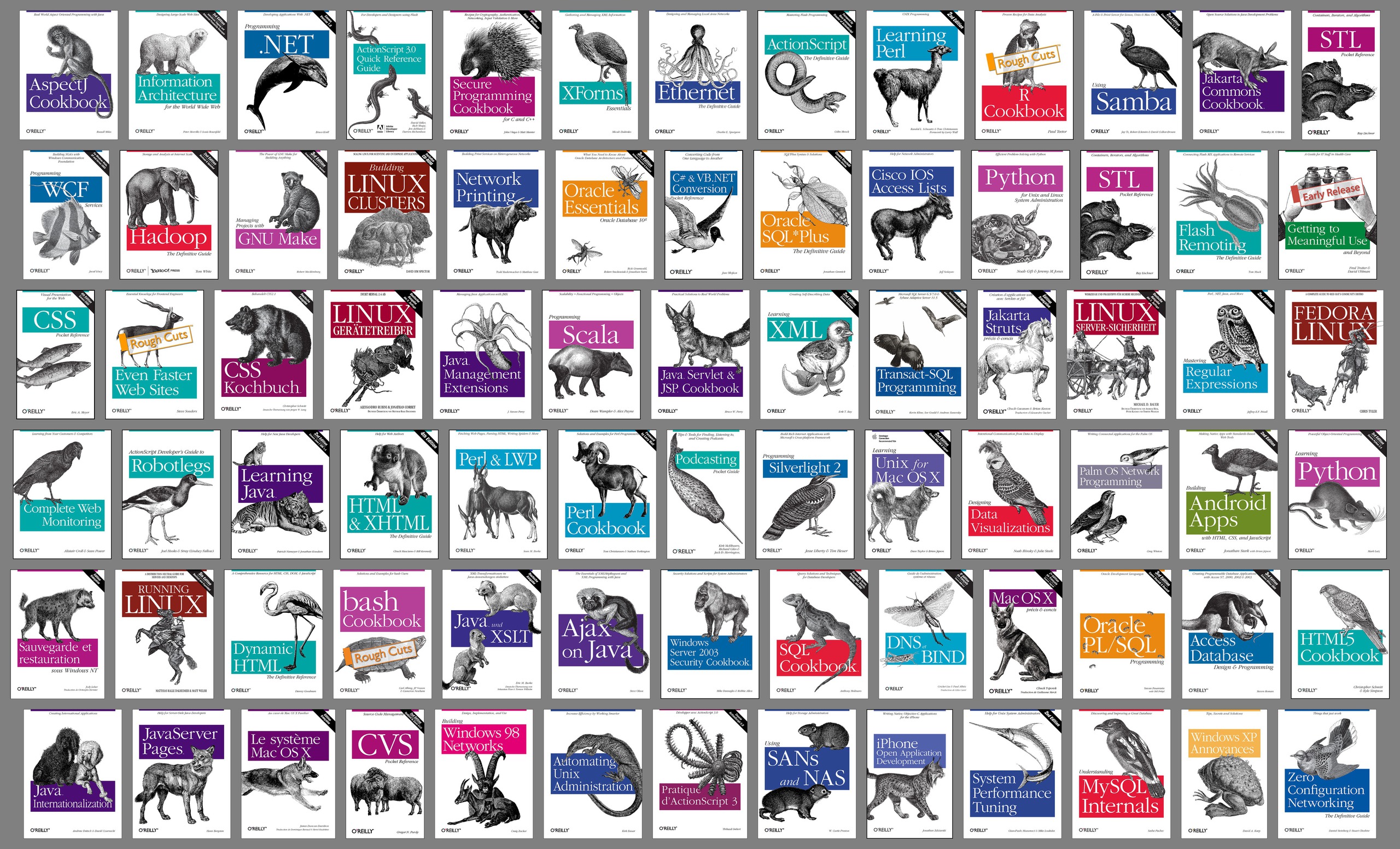}\\[0.1cm]
\includegraphics[width=\columnwidth]{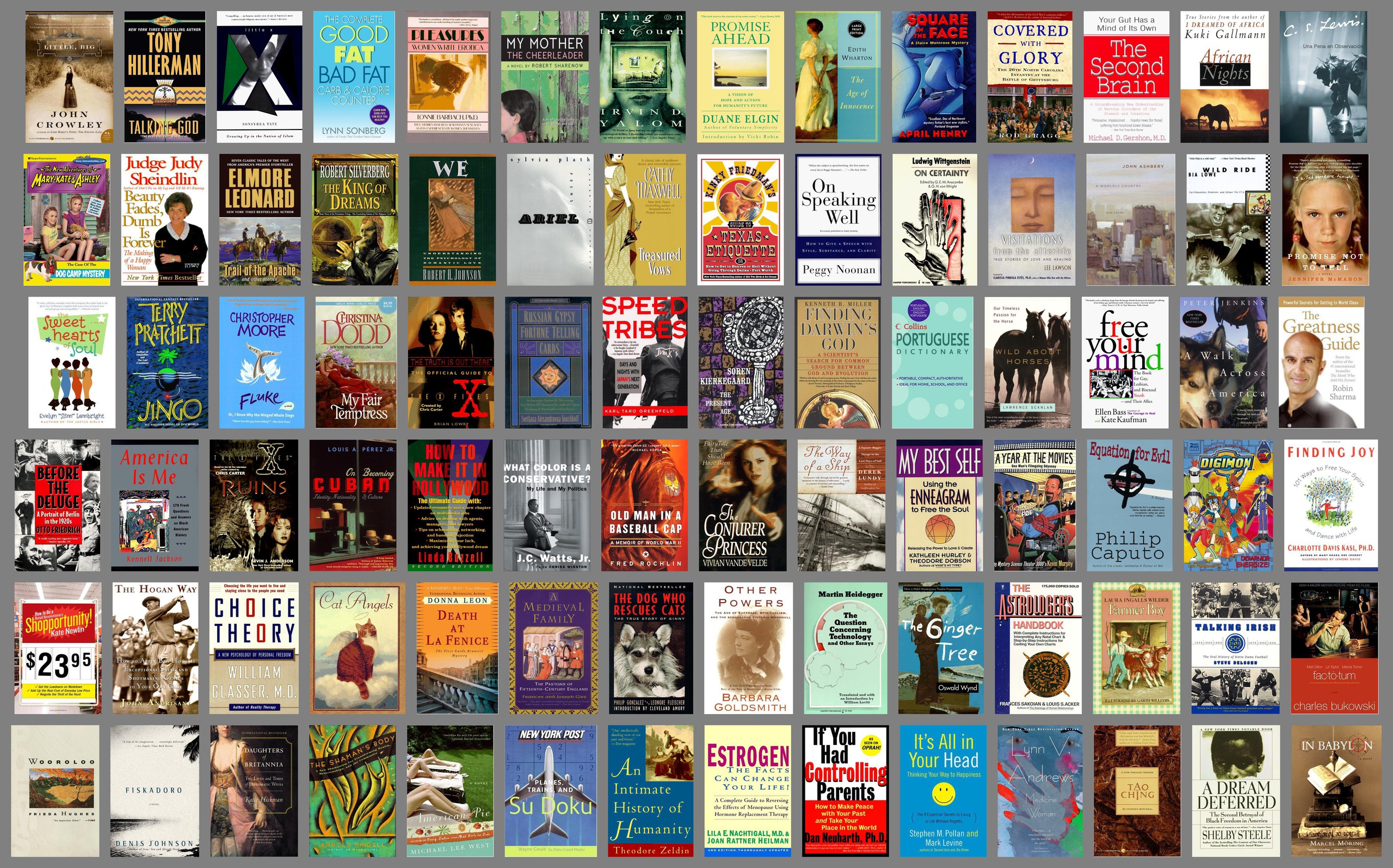}\\[0.1cm]
\includegraphics[width=\columnwidth]{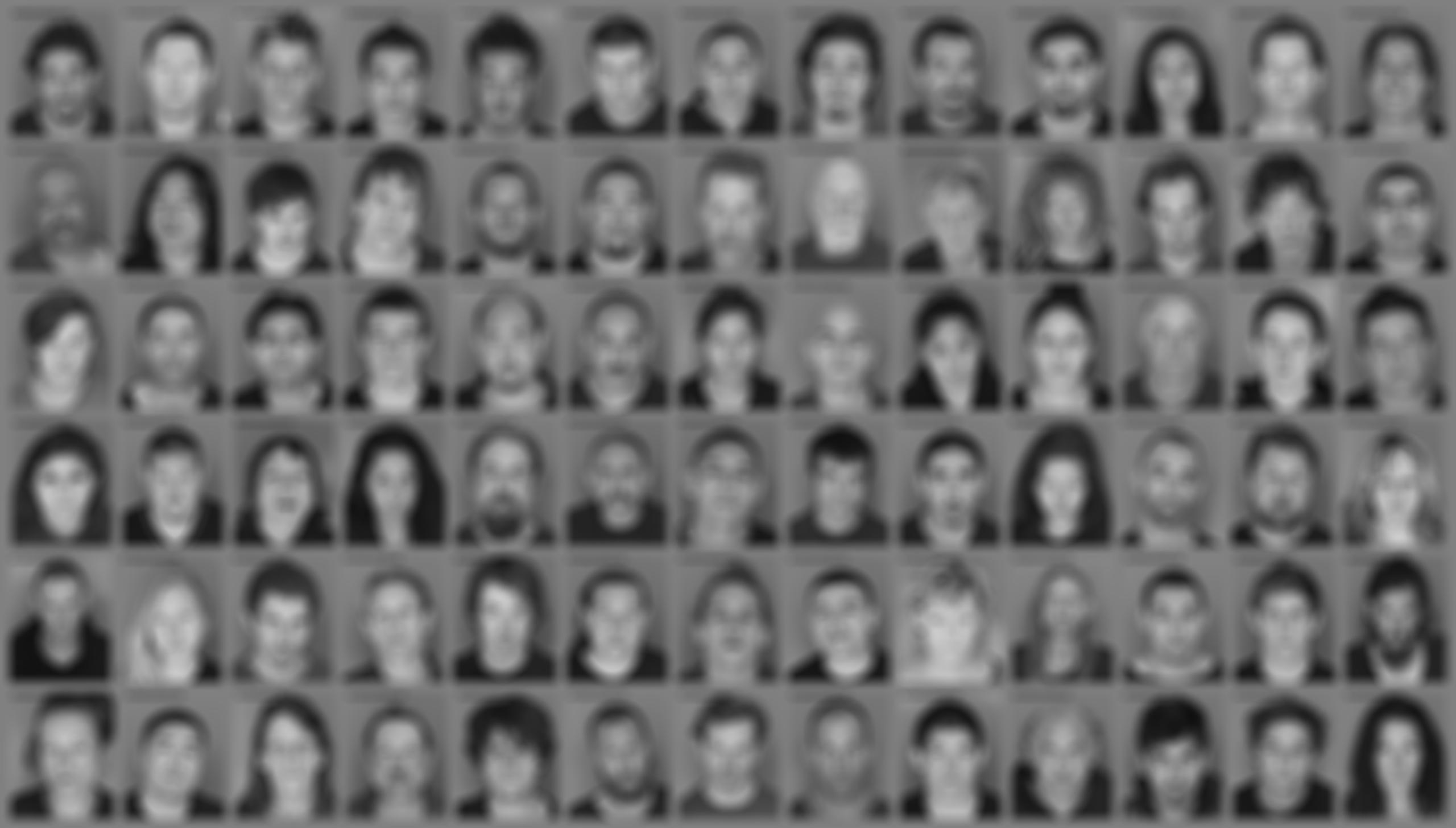}
\caption{Sample image collages used for data collection: O'Reilly book covers (top), Amazon book covers (middle), mugshots (bottom, blurred for privacy reasons). Participants were asked to find different targets within random permutations of these collages.}
\label{fig:collages}
\end{figure}

Given the lack of an appropriate dataset, we designed a human study to collect fixation data during visual search. In contrast to previous works that used $3 \times 3$ squared patterns at two or three luminance levels, or synthesised images of natural scenes~\cite{borji2014eyes}, our goal was to collect fixations on collages of natural images. 
We therefore opted for a task that involved searching for a single image (the target) within a synthesised collage of images (the search set). Each of the collages are the random permutation of a finite set of images. To explore the impact of the similarity in appearance between target and search set on both fixation behaviour and automatic inference, we have created three different search tasks covering a range of similarities. 

In prior work, colour was found to be a particularly important cue for guiding search to targets and target-similar objects~\cite{hwang2009model,motter1998guidance}.
Therfore we have selected for the first task 78 coloured O'Reilly book covers to compose the collages. These covers show a woodcut of an animal at the top and the title of the book in a characteristic font underneath (see Figure~\ref{fig:collages} top). Given that overall cover appearance was very similar, this task allows us to analyse fixation 
behaviour when colour is the most discriminative feature.

For the second task we use a set of 84 book covers from Amazon. In contrast to the first task, appearance of these covers is more diverse (see Figure~\ref{fig:collages} middle). This makes it possible to analyse fixation behaviour when both structure and colour information could be used by participants to find the target.

Finally, for the third task, we use a set of 78 mugshots from a public database of suspects.
In contrast to the other tasks, we transformed the mugshots to grey-scale so that they did not contain any colour information (see Figure~\ref{fig:collages} bottom). In this case, allows abalysis of fixation behaviour when colour information was not available at all. We found faces to be particularly interesting given the relevance of searching for faces in many practical applications.

We place images on a grid in order to form collages that we show to the participants.
Each collage is a random permutation of the available set of images on the grid. The search targets are subset of images in the collages. 
We opted for an independent measures design to reduce fatigue (the current recording already took 30 minutes of concentrated search to complete) and learning effects that both may have influenced fixation behaviour.

\subsection{Participants, Apparatus, and Procedure}

We recorded fixation data of 18 participants (nine male) with different nationalities and aged between 18 and 30 years. The eyesight of nine participants was impaired but corrected with contact lenses or glasses. 
To record gaze data we used a stationary Tobii TX300 eye tracker that provides binocular gaze data at a sampling frequency of 300Hz. Parameters for fixation detection were left at their defaults: fixation duration was set to 60ms while the maximum time between fixations was set to 75ms. The stimuli were shown on a 30 inch screen with a resolution of 2560x1600 pixels.

Participants were randomly assigned to search for targets for one of the three stimulus types.
We first calibrated the eye tracker using a standard 9-point calibration, followed by a validation of eye tracker accuracy. 
After calibration, participants were shown the first out of five search targets. Participants had a maximum of 10 seconds to memorise the image and 20 seconds to subsequently find the image in the collage. Collages were displayed full screen and consisted of a fixed set of randomly ordered images on a grid.
The target image always appeared only once in the collage at a random location.

To determine more easily which images participants fixated on, all images were placed on a grey background and had a margin to neighbouring images of on average 18 pixels.
As soon as participants found the target image they pressed a key. Afterwards they were asked whether they had found the target and how difficult the search had been. This procedure was repeated twenty times for five different targets, resulting in a total of 100 search tasks.
To minimise lingering on search taget, participants were put under time pressure and had to find the target and press a confirmation button as quickly as possible.
This resulted in lingering of $2.45\%$ for Amazon (O'Reilly: $1.2\%$, mugshots: $0.35\%$).

\section{Method}

In this work we are interested in search tasks in which the fixation patterns are modulated by the search target. Previous work focused on predicting a fixed set of targets for which fixation data was provided at training time. We call this the {\it closed-world setting}. In contrast, our method enables prediction of new search targets, i.e. those for which no fixation is available for training. We refer to this as the {\it open-world setting}. 
In the following, we first provide a problem formulation for the previously investigated {\it closed-world setting}. Afterwards we present a new problem formulation for search target prediction in an {\it open-world setting} (see Figure \ref{fig:evaluations}).

\subsection{Search Target Prediction}

Given a query image (search target) $Q \in \mathbb{Q}$ and a stimulus collage $C \in \mathbb{C}$, during a search task participants $P \in \mathbb{P}$ perform fixations $F(C,Q,P) = \{(x_i,y_i,a_i), i=1,\dots,N\}$, where each fixation is a triplet of positions $x_i,y_i$ in screen coordinates and appearance $a_i$ at the fixated location.
To recognise search targets we aim to find a mapping from fixations to query images:

\begin{equation}
F(C,Q) \mapsto Q \in \mathbb{Q}
\end{equation}

We use a bag of visual world featurisation $\phi$ of the fixations. We interpret fixations as key points around which we extract local image patches. These are clustered into a visual vocabulary $V$ and accumulated in a count histogram. This leads to a fixed-length vector representation of dimension $|V|$ commonly known as a bag of words. Therefore, our recognition problem can more specifically be expressed as:

\begin{equation}
\phi(F(C,Q,P), V) \mapsto Q \in \mathbb{Q}
\end{equation}

\subsection{Closed-World Setting}

We now formulate the previously investigated case of the closed-world setting where all test queries (search targets) $Q \in \mathbb{Q}_\text{test}$ are part of our training set $\mathbb{Q}_\text{test} = \mathbb{Q}_\text{train}$ and, in particular, we assume that we observe fixations $\mathbb{F}_\text{train} = \{F(C,Q,P) | \forall Q \in \mathbb{Q}_\text{train}\}$. 
The task is to predict the search target while the query and/or participant changes (see Figure~\ref{fig:evaluations}).

\begin{equation}
\phi(F(C,Q,P), V) \mapsto Q \in \mathbb{Q}_\text{train}
\end{equation}

We use a one-vs-all multi-class SVM classifier $\mathcal{H}_i$ and the query image with the largest margin:
\begin{equation}
Q_i=\argmax_{i=1,\dots,|\mathbb{Q}_\text{test}|} \mathcal{H}_{i}(\phi(F_\text{test},V) ) \; 
\end{equation}

\subsection{Open-World Setting}

In contrast, in our new open-world setting, we no longer assume that we have fixation data to train for these targets. Therefore $\mathbb{Q}_\text{test} \cap \mathbb{Q}_\text{train} = \emptyset$. The main challenge that arises from this setting is to develop a learning mechanism that can predict over a set of classes that is unknown at training time (see Figure~\ref{fig:evaluations}). 

\paragraph{Search Target Prediction}

To circumvent the problem of training for a fixed number of search targets, we propose to encode the search target into the feature vector, rather than considering it a class that is to be recognised. This leads to a formulation where we learn compatibilities between observed fixations and query images:
\begin{equation}
(F(C,Q_i,P),Q_j)
\mapsto
Y \in \{0,1\}
\end{equation}

Training is performed by generating data points of all pairs of $Q_i$ and $Q_j$ in $\mathbb{Q}_\text{train}$ and assigning a compatibility label $Y$ accordingly:
\begin{equation}
Y = \left\{
  \begin{array}{l l}
    1 & \quad \text{if $i=j$ }\\
    0& \quad \text{if $i\neq j$}
  \end{array} \right.
  \end{equation}
The intuition behind this approach is that the compatibility predictor learns about similarities in fixations and search targets that can also be applied to new fixations and search targets.

Similar to the closed-world setting, we propose a featurisation of the fixations and query images. Although we can use the same fixation representation as before, we do not have fixations for the query images. Therefore, we introduce a sampling strategy $S$ which still allows us to generate a bag-of-words representation for a given query. In this work we propose to use sampling from the saliency map as a sampling strategy. We stack the representation of the fixation and the query images. This leads to the following learning problem:
\begin{equation}
\left(
\begin{array}{c}
\phi(F(C,Q_i,P),V)\\
\phi(S(Q_j))
\end{array}
\right)
\mapsto
Y \in \{0,1\}
\end{equation}
We learn a model for the problem by training a single binary SVM $\mathcal{B}$ classifier according to the labelling as described above. At test time we find the query image describing the search target by
\begin{equation}
Q = \argmax_{Q_j \in \mathbb{Q}_\text{test}} \mathcal{B} 
\left(
\begin{array}{c}
\phi(F_\text{test},V)\\
\phi(S(Q_j))
\end{array}
\right)
\end{equation}

Note that while we do not require fixation data for the query images that we want to predict at test time, we still search over a finite set of query images $\mathbb{Q}_\text{test}$.

\section{Experiments}

\begin{figure}[t]
\centering
\includegraphics[width=0.5\columnwidth]{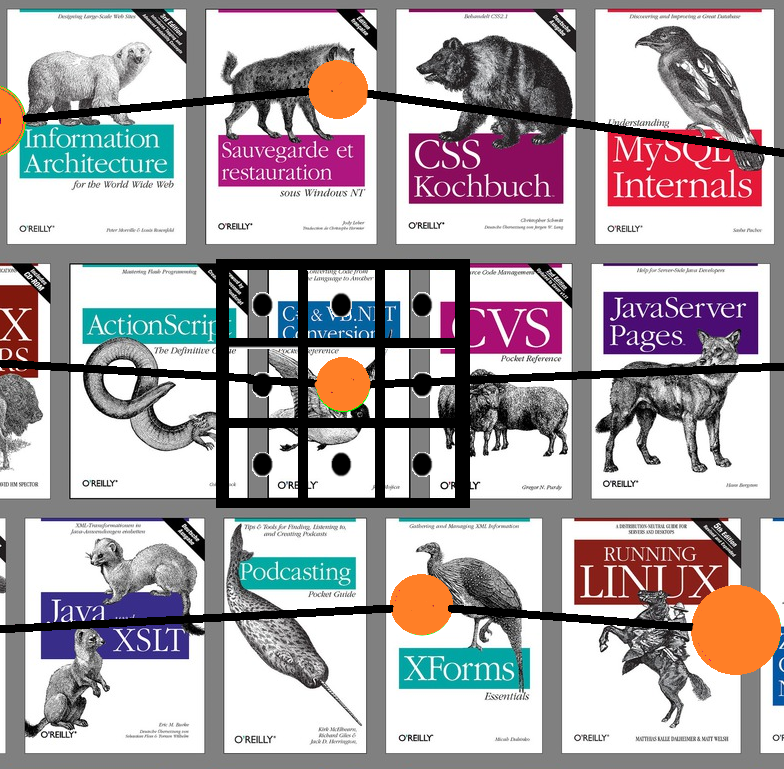}
\caption{Proposed approach of sampling eight additional image patches around each fixation location to compensate for eye tracker inaccuracy. The size of orange dots corresponds to the fixation's duration.}
\label{fig:patch}
\end{figure}
 
Our dataset contains fixation data from six participants for each search task. To analyse the first and second search task
(O'Reilly and Amazon book covers) we used RGB values extracted from a patch (window) of size $m\times m$ around each fixation as input to the bag-of-words model. 
For the third search task (mugshots) we calculated a histogram of local binary patterns from each fixation patch.
To compensate for inaccuracies of the eye tracker we extracted eight additional points with non-overlapping patches around each fixation (see Figure~\ref{fig:patch}).
Additionally, whenever an image patch around a fixation had overlap with two images in the collage, pixel values in the area of the overlap were set to 128.

\subsection{Closed-World Evaluation}

In our closed-world evaluation we distinguish between within-participant and cross-participant predictions.
In the ``within participant'' condition we predict the search target for each participant individually using their own training data.
In contrast, for the ``cross participant'' condition, we predict the search target across participants. The ``cross participant'' condition is more challenging as the algorithm has to generalise across users.
Chance level is defined based on the number of search targets or classes our algorithm is going to predict. Participants were asked to search for five different targets in each experiment (chance level $1/5 = 20\%$).

\subsubsection*{Within-Participant Prediction}

\begin{figure}[t]
\centering
\includegraphics[width=1\columnwidth]{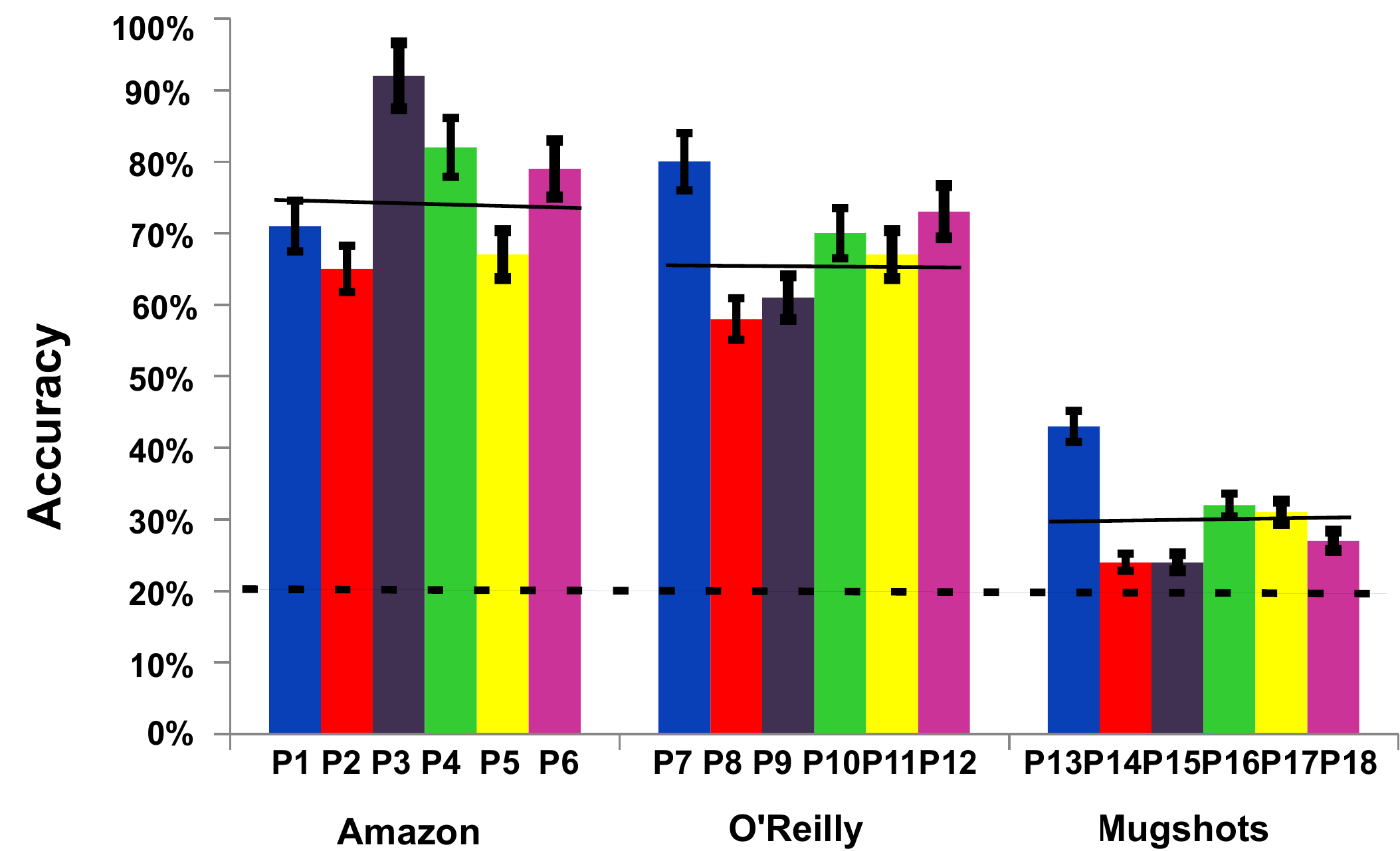}
\caption{Closed-world evaluation results showing mean and standard deviation of within-participant prediction accuracy for Amazon book covers, O'Reilly book covers, and mugshots. Mean performance is indicated with black lines, and the chance level is indicated with the dashed line.}
\label{fig:Intraperson_optimum}
\end{figure}

Participants looked for each search target 20 times. To train our classifier we used the data from 10 trials and the remaining 10 trials were used for testing.
We fixed the patch (window) size to $41\times 41$ and optimised $k$ (vocabulary size) for each participant. Figure~\ref{fig:Intraperson_optimum} summarises the within-participant prediction accuracies for the three search tasks.
Accuracies were well above chance for all participants for the Amazon book covers (average accuracy $75\%$) and the O'Reilly book covers (average accuracy $69\%$). 
Accuracies were lower for mugshots but still above chance level (average accuracy $30\%$, chance level $20\%$). 

\subsubsection*{Cross-Participant Prediction}

We investigated whether search targets could be predicted within and across participants. In the accross-participants case, we trained one-vs-all multi-class SVM classifier using 3-fold cross-validation. We trained our model with data from three participants to map the observer-fixated patch to the target image. The resulting classifier was then tested on data from the remaining three participants. Prior to our experiments, we ran a control experiment where we uniformly sampled from $75\%$ of the salient part of the collages.
We trained the classifier with these randomly sampled fixations and confirmed that performance was around the chance level of 20\% and therefore any improvement can indeed be attributed to information contained in the fixation patterns.

\begin{figure}[t]
        \centering
        \begin{subfigure}[b]{0.9\columnwidth}
                \includegraphics[width=\textwidth]{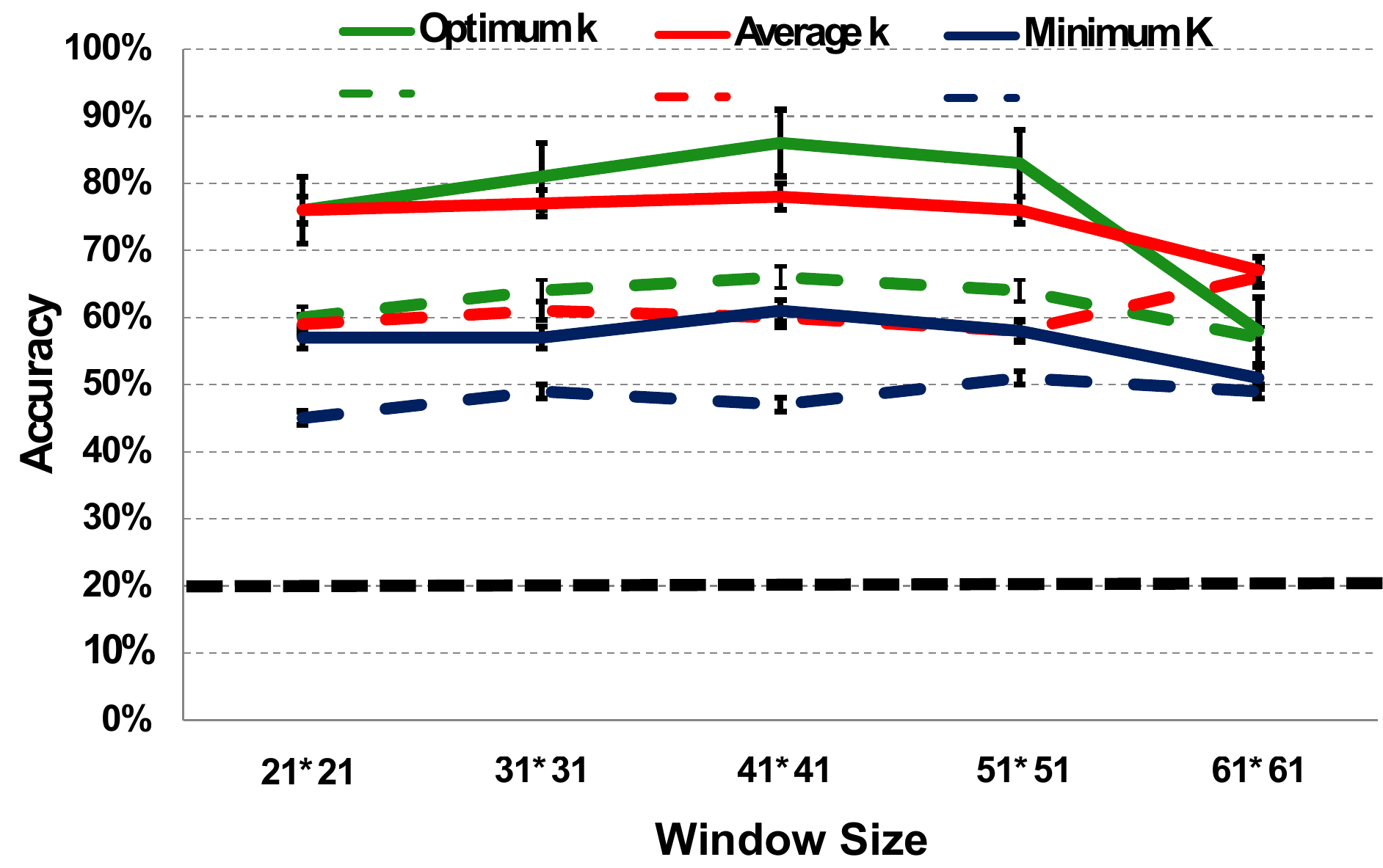}
                \label{fig:inter:usersamplesnumber}
        \end{subfigure}

        \begin{subfigure}[b]{0.9\columnwidth}
                \includegraphics[width=\textwidth]{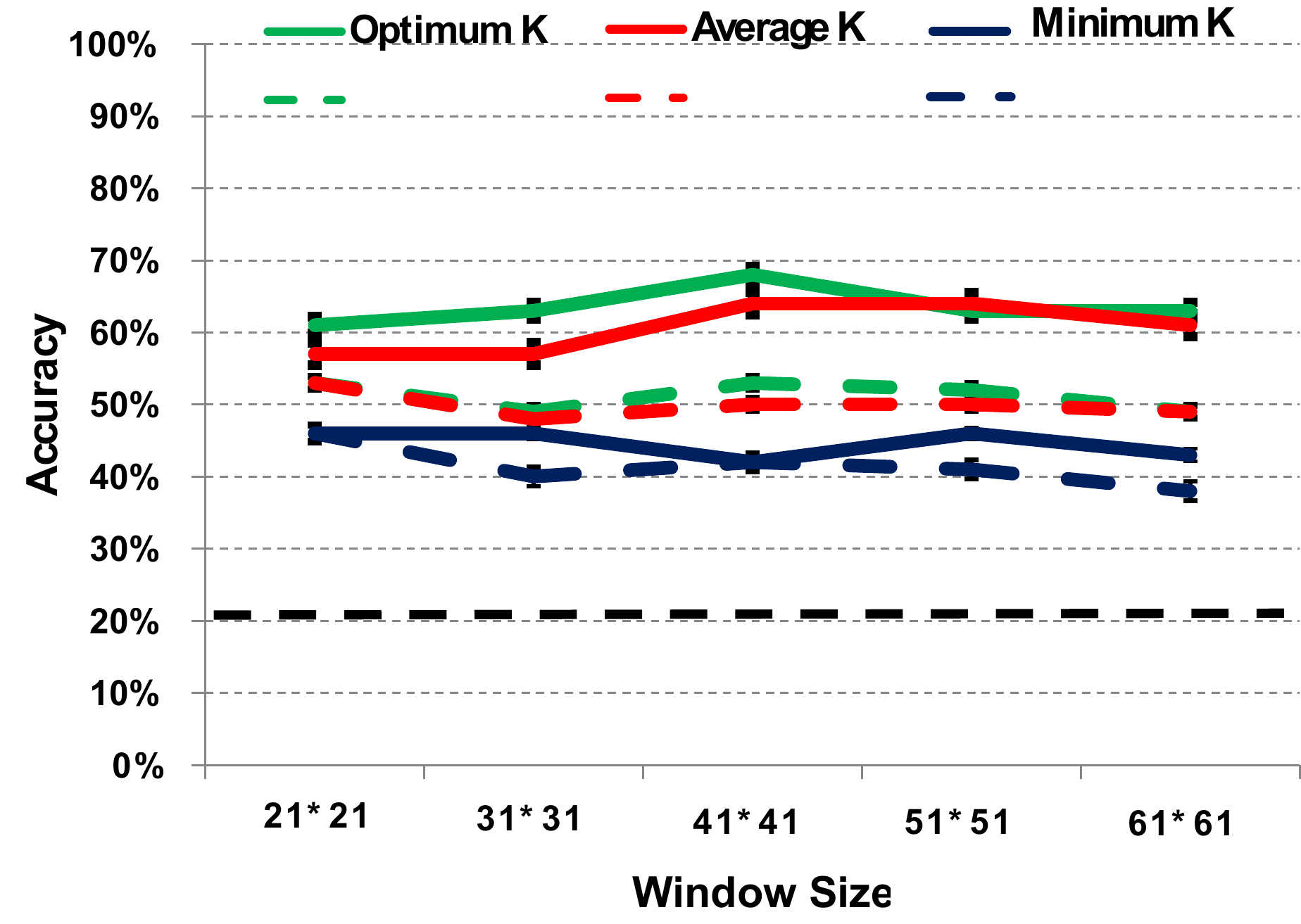}
                \label{fig:inter:timeofday}
        \end{subfigure}

        \begin{subfigure}[b]{0.9\columnwidth}
                \includegraphics[width=\textwidth]{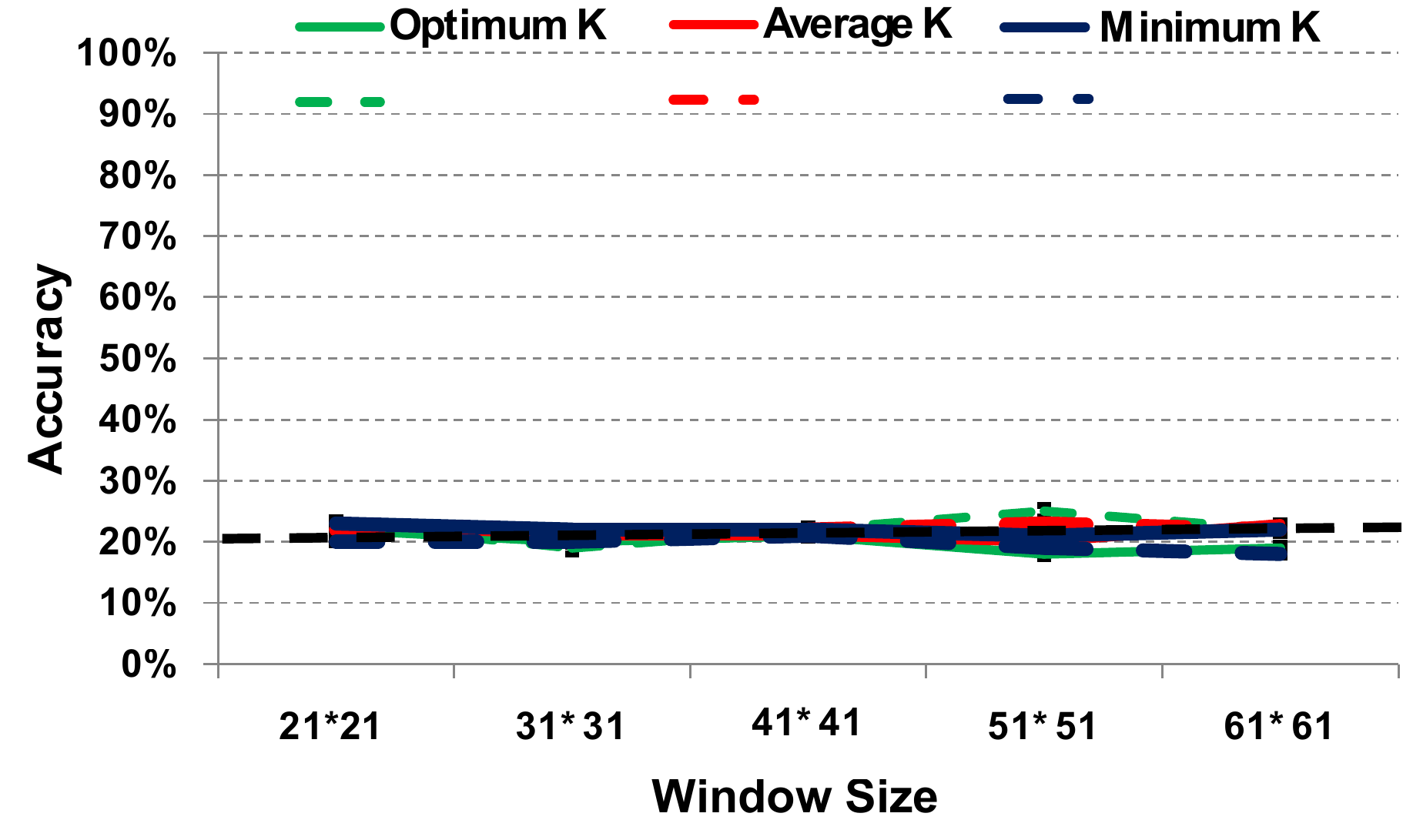}
                \label{fig:inter:intensity}
        \end{subfigure}
        \caption{Closed-world evaluation results showing mean and standard deviation of cross-participant prediction accuracy for Amazon book covers (top), O'Reilly book covers (middle), and mugshots (bottom). Results are shown with (straight lines) and without (dashed lines) using the proposed sampling approach around fixation locations. The chance level is indicated with the dashed line.}
    \label{fig:inter}
\end{figure}

Figure~\ref{fig:inter} summarises the cross-participant prediction accuracies for Amazon book covers, O'Reilly book covers, and mugshots for different window sizes and size of vocabulary $k$, as well as results with (straight lines) and without (dashed lines) using the proposed sampling approach around fixation locations. The optimum $k$ represents the upper bound and corresponds to always choosing the value of $k$ that optimises accuracy, while the minimum $k$ correspondingly represents the lower bound. Average $k$ refers to the practically most realistic setting in which we fix $k=60$. Performance for Amazon book covers was best, followed by O'Reilly book covers and mugshots. Accuracies were between $61\%\pm2\%$ and $78\%\pm2\%$ for average $k$ for Amazon and O'Reilly book covers but only around chance level for mugshots.

\subsection{Open-World Evaluation}

In the open-world evaluation the challenge is to predict the search target based on the similarity between fixations $F(C,Q)$ and query image $S(Q)$.
In absence of fixations for query images $Q$ we uniformly sample from the GBVS saliency map~\cite{harel2006graph}.
We chose the number of samples on the same order as the number of fixations on the collages.
For the within-participant evaluation we used the data from three search targets of each participant to train a binary SVM with RBF kernel.
The data from the remaining two search targets was used at test time.
The average performance of all participants in each group was for Amazon: $70.33\%$, O'Reilly: $59.66\%$, mugshots: $50.83\%$.

Because the task is more challenging in the cross-participant evaluation, we report results for this task in more detail. 
As described perviously, we train a binary SVM with RBF kernel from data of three participants to learn the similarity between the observer-fixated patch when looking for three of the search targets and the corresponding target images.
Our positive class contains data coming from the concatenation of $\Phi(F(C,Q_{i},P),V)$ and $\Phi(S(Q_{j}))$ when $i=j$. At test time, we then test on the data of remaining three participants looking for two other search targets that did not appear in the training set and the corresponding search targets.
The chance level is $1/2=50\%$ as we have a target vs, non-target decision.

\begin{figure}[t]
        \centering
        \begin{subfigure}[b]{0.77\columnwidth}
                \includegraphics[width=\textwidth]{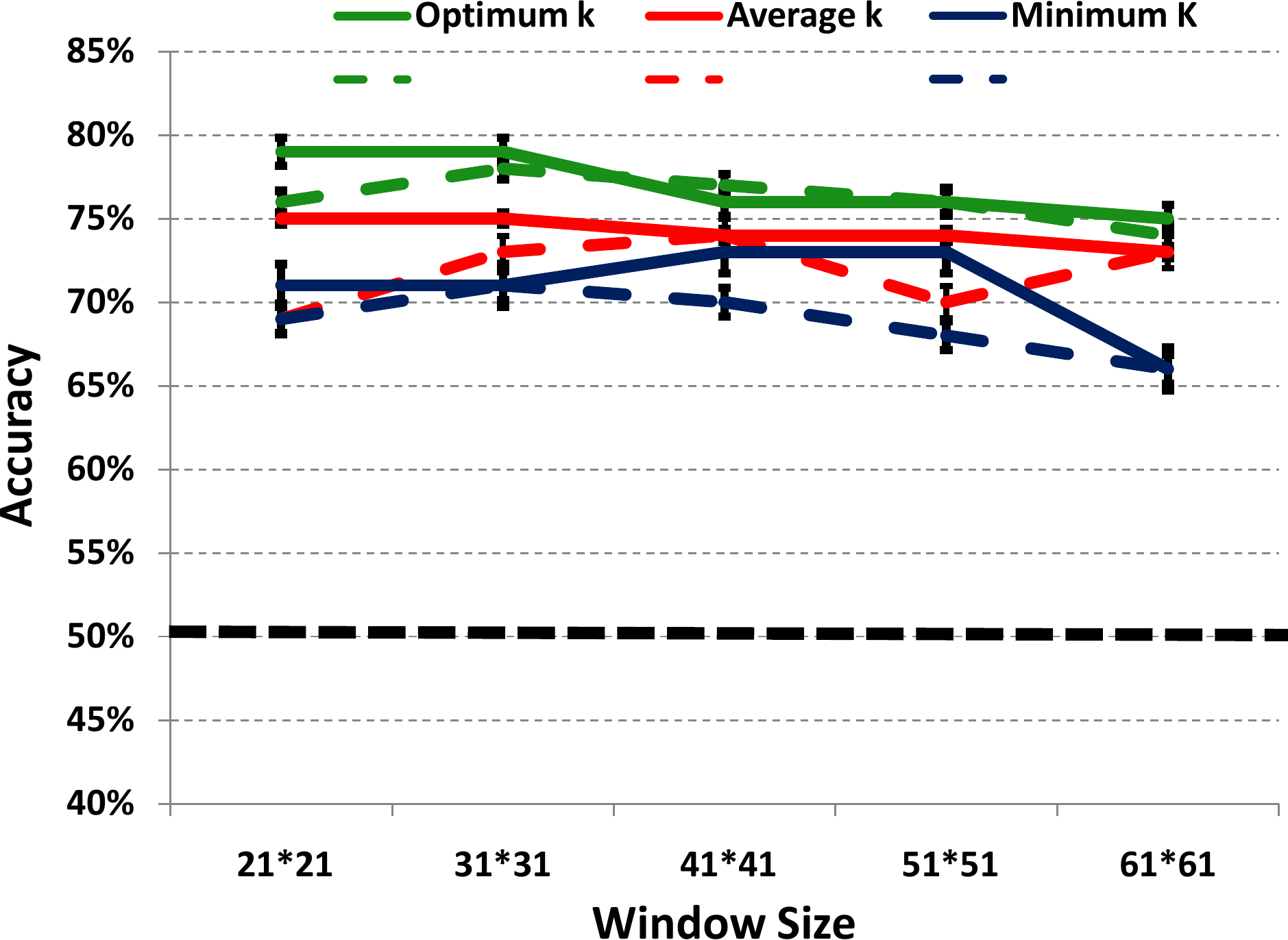}
                \label{fig:open:usersamplesnumber}
        \end{subfigure}

        \begin{subfigure}[b]{0.77\columnwidth}
                \includegraphics[width=\textwidth]{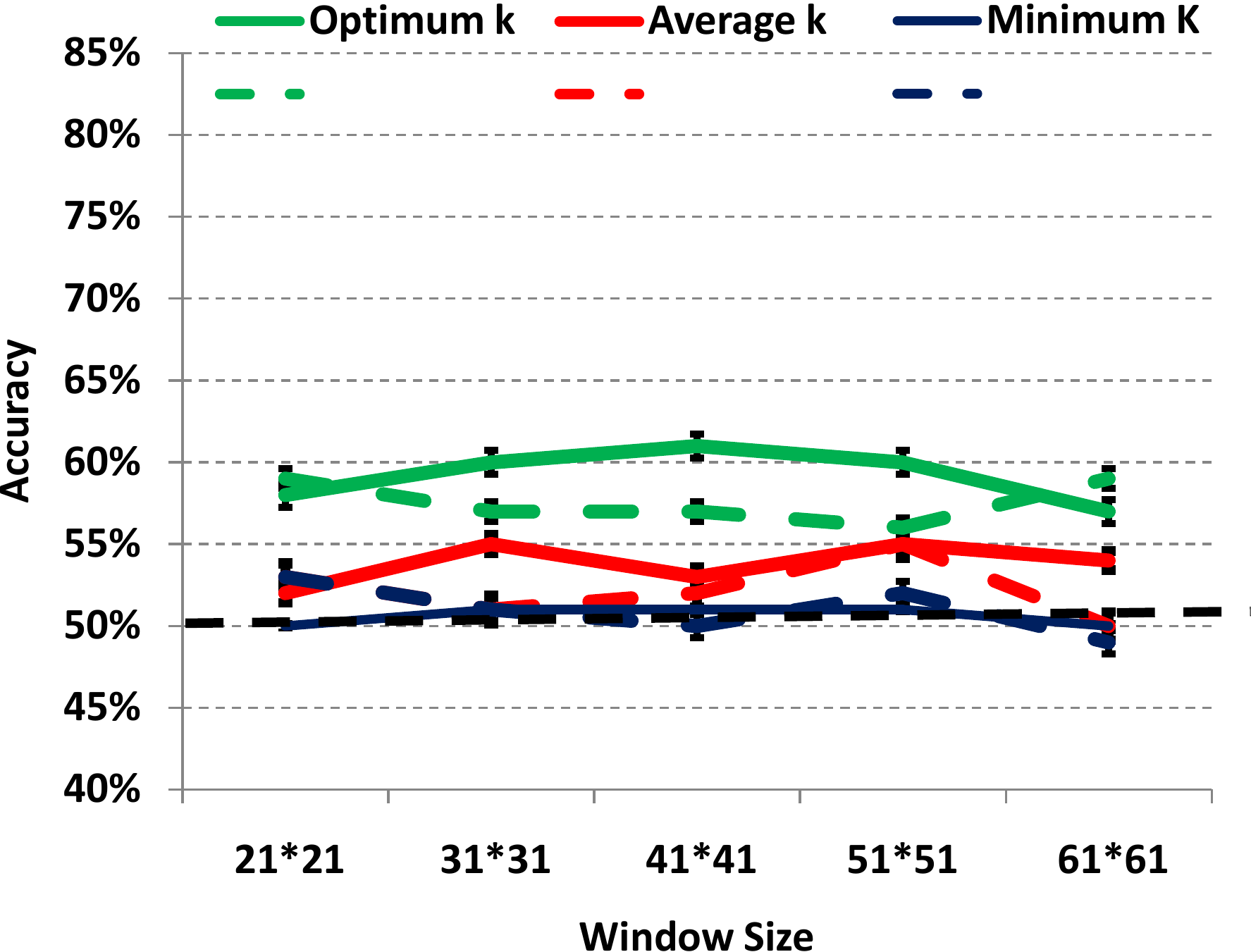}
                \label{fig:open:timeofday}
        \end{subfigure}

        \begin{subfigure}[b]{0.77\columnwidth}
                \includegraphics[width=\textwidth]{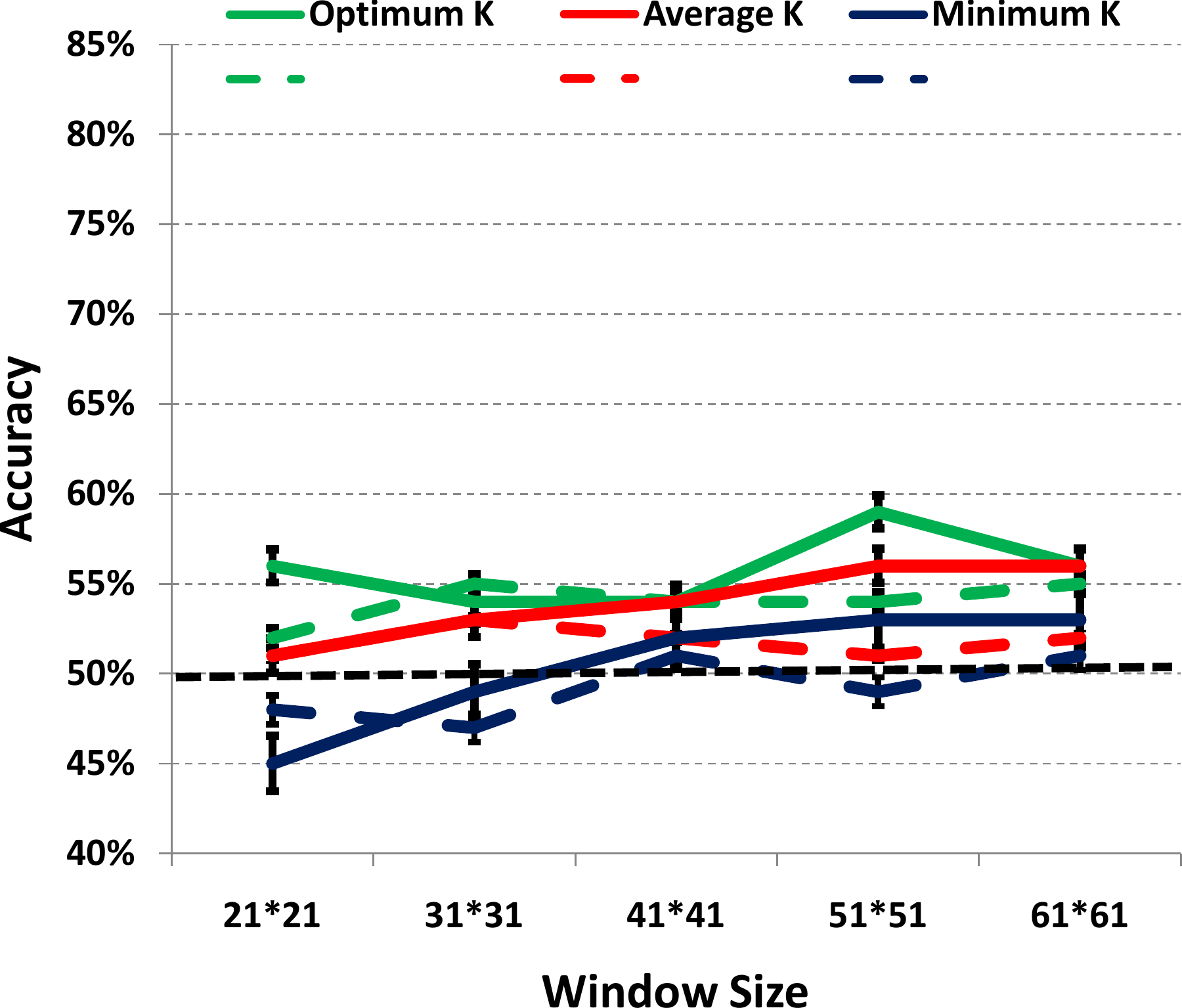}
                \label{fig:open:intensity}
        \end{subfigure}
        \caption{Open-World evaluation results showing mean and standard deviation of cross-participant prediction accuracy for Amazon book covers (top), O'Reilly book covers (middle), and mugshots (bottom). Results are shown with (straight lines) and without (dashed lines) using the proposed sampling approach around fixation locations. The chance level is indicated with the dashed line.}
    \label{fig:open}
\end{figure}

Figure~\ref{fig:open} summarises the cross-participant prediction accuracies for Amazon book covers, O'Reilly book covers, and mugshots for different window sizes and size of vocabulary $k$, as well as results with (straight lines) and without (dashed lines) using the proposed sampling approach around fixation locations. With average $k$ the model achieves an accuracy of $75\%$ for Amazon book covers, which is significantly higher than chance at $50\%$. For O'Reilly book covers accuracy reaches $55\%$ and for mugshots we reach $56\%$. 
Similar to our closed-world setting, accuracy is generally better when using the proposed sampling approach.

\begin{figure}[t]
\centering
\includegraphics[width=1\columnwidth]{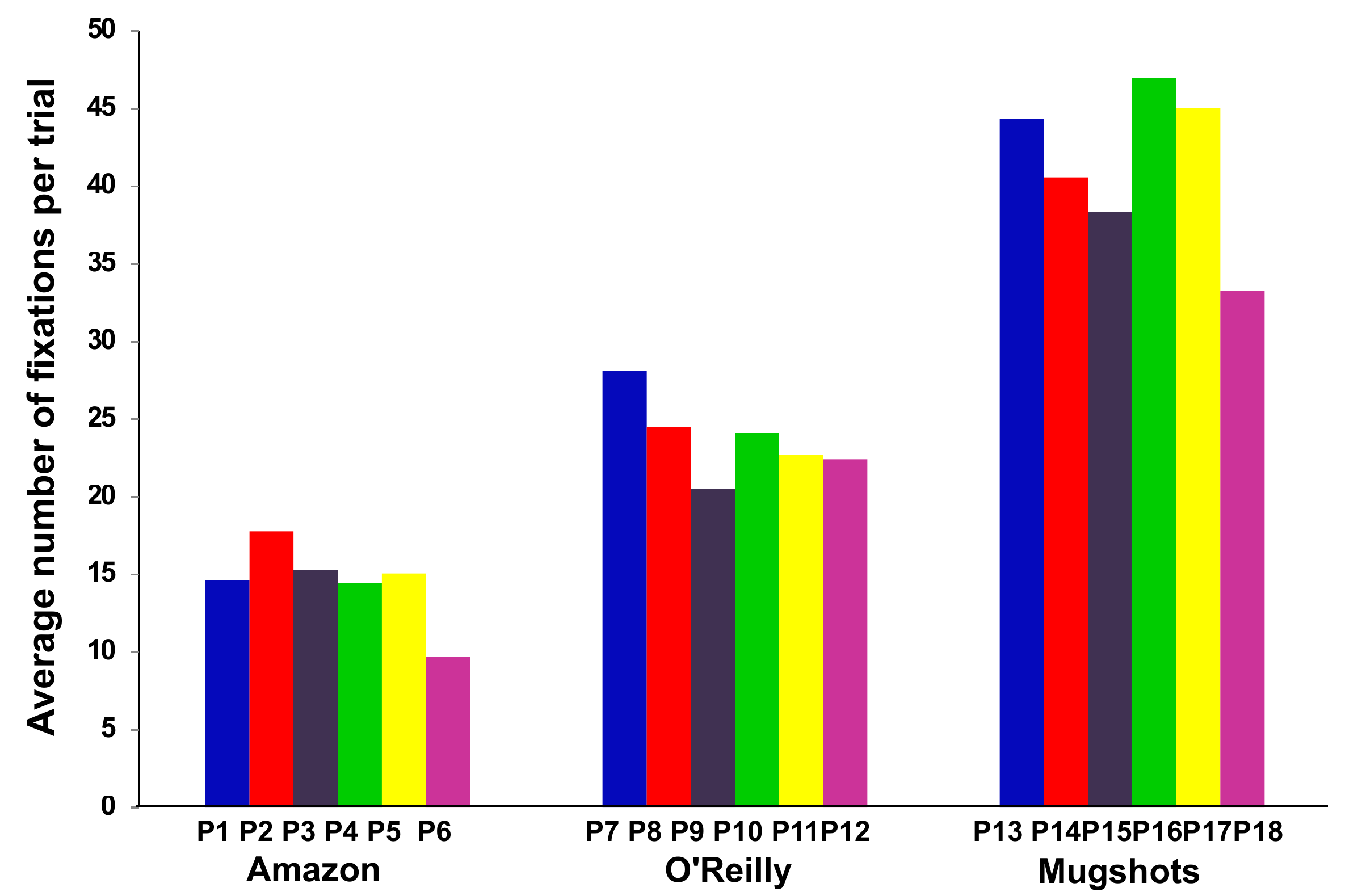}
\caption{Average number of fixations per trial performed by each participant during the different search tasks.}
\label{fig:numfix}
\end{figure}

\section{Discussion}

In this work we studied the problem of predicting the search target during visual search from human fixations. 
Figure~\ref{fig:Intraperson_optimum} shows that we can predict the search target significantly above chance level for the within-participant case for the Amazon and O'Reilly book cover search tasks, with accuracies ranging from $50\%$ to $78\%$. Figure~\ref{fig:inter} shows similar results for the cross-participant case. These findings are in line with previous works on search target prediction in closed-world settings~\cite{zelinsky2013eye,borji2014eyes}. Our findings extend these previous works in that we study synthesised collages of natural images and in that our method has to handle a larger number of distractors, higher similarities between search image and distractors, and a larger number of potential search targets. Instead of a large number of features, we rely only on colour information as well as local binary  pattern features.

\begin{figure}[t]
\centering
\includegraphics[width=1\columnwidth]{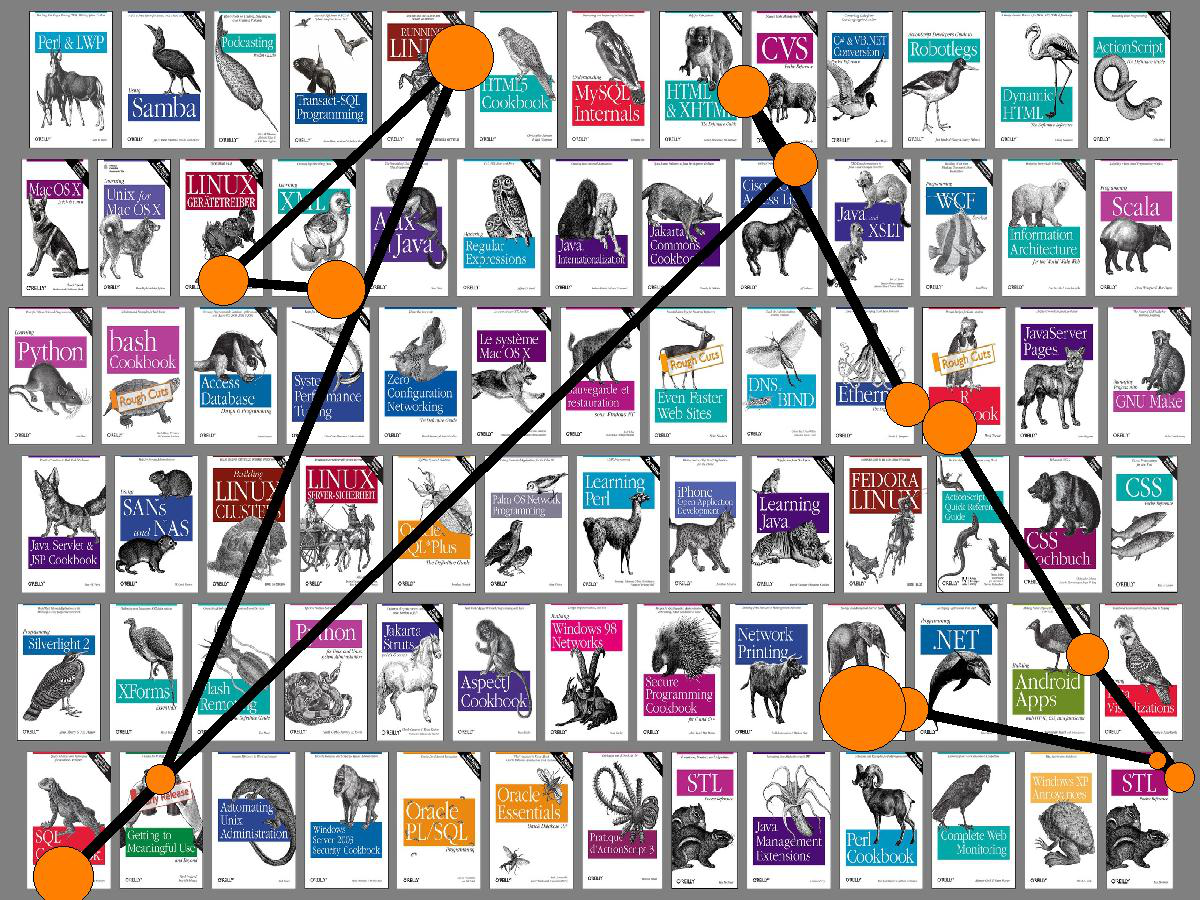}\\[0.1cm]
\includegraphics[width=1\columnwidth]{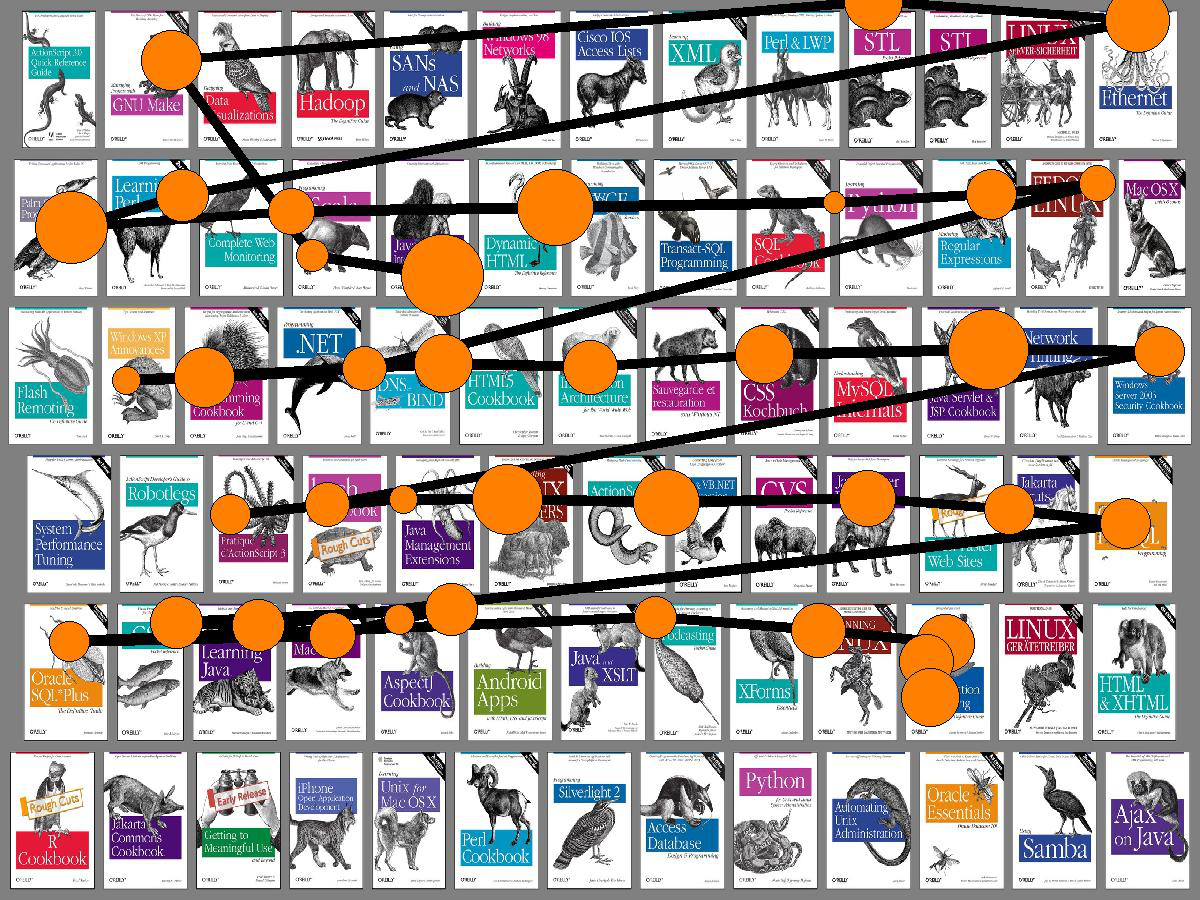}
\caption{Sample scanpaths of P8: Targeted search behaviour with a low number of fixations (top), and skimming behaviour with a high number of fixations (bottom). Size of the orange dots corresponds to fixation durations.}
\label{fig:taskdiff}
\end{figure}

We extended these evaluations with a novel open-world evaluation setting in which we no longer assume that we have fixation data to train for these targets. To learn under such a regime we proposed a new formulation where we learn compatibilities between observed fixations and query images. As can be seen from Figure~\ref{fig:open}, despite the much more challenging setting, using this formulation we can still predict the search target significantly above chance level for the Amazon book cover search task, and just about chance level for the other two search tasks for selected values of $k$. These results are meaningful as they underline the significant information content available in human fixation patterns during visual search, even in a challenging open-world setting. The proposed method of sampling  eight additional image patches around each fixation to compensate for eye tracker inaccuracies proved to be necessary and 
effective for both evaluation settings and increased performance in the closed-world setting by up to 20\%, and by up to 5\% in the open-world setting.

These results also support our initial hypothesis that the search task, i.e.\ in particular the similarity in appearance between target and search set and thus the difficulty, has a significant impact on both fixation behaviour and prediction performance. Figures~\ref{fig:inter} and~\ref{fig:open} show that we achieved the best performance for the Amazon book covers, for which appearance is very diverse and participants can rely on both structure and colour information. The O'Reilly book covers, for which the cover structure was similar and colour was the most discriminative feature, achieved the second best performance. In contrast, the worst performance was achieved for the greyscale mugshots that had highly similar structure and did not contain any colour information. These findings are in line with previous works in human vision that found that colour is a particularly important cue for guiding search to targets and target-similar objects~\cite{motter1998guidance,hwang2009model}.

\begin{figure}[t]
\centering
\includegraphics[width=1\columnwidth]{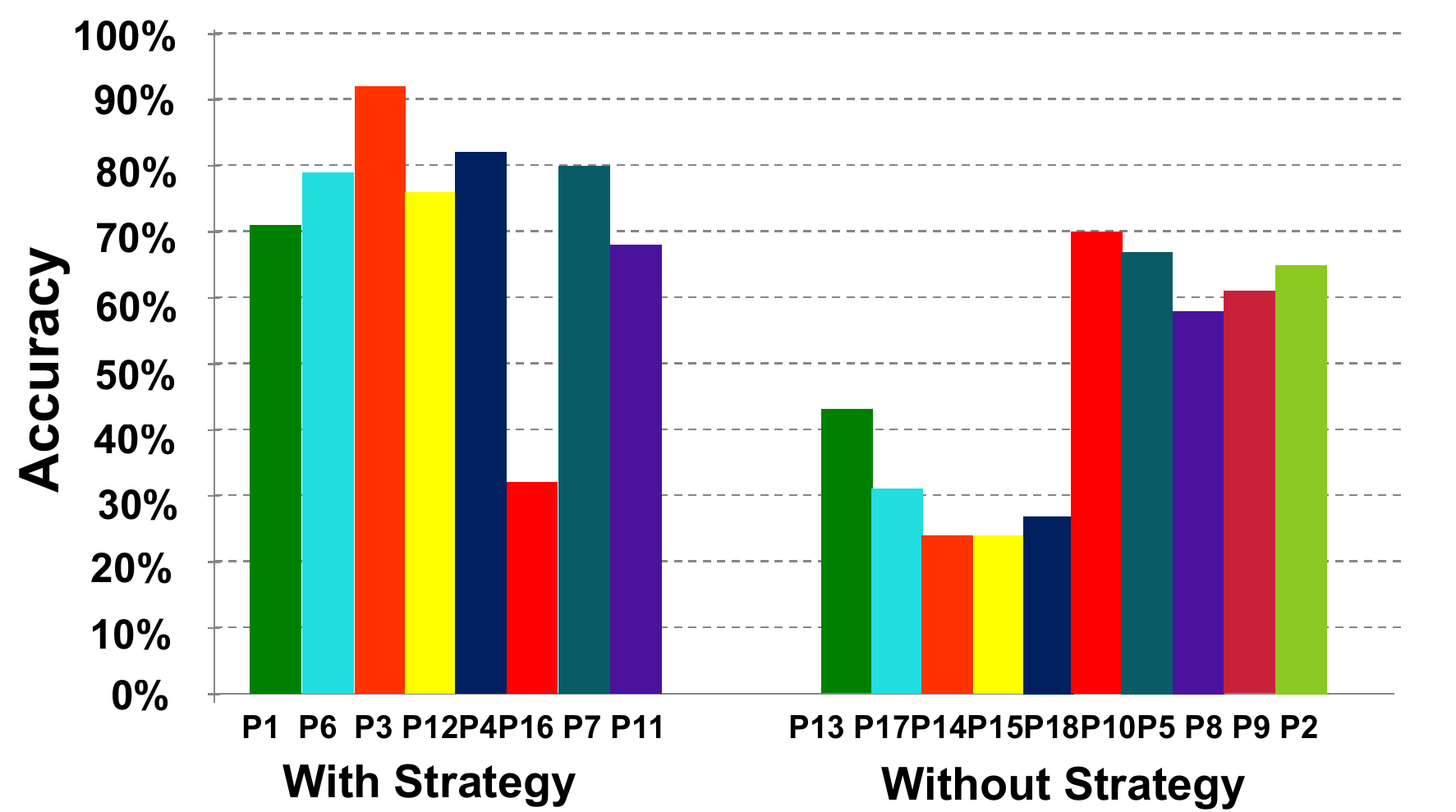}
\caption{Difference in accuracies of participants who have a strategic search pattern vs participants that mainly skim the collage to find the search image.}
\label{fig:ac-skim}
\end{figure}

Analysing the visual strategies that participants used provides additional interesting (yet anecdotal) insights.
As the difficulty of the search task increased, participants tended to start skimming the whole collage 
rather than doing targeted search for specific visual features (see Figure~\ref{fig:taskdiff} for an example). This tendency was the strongest for the most difficult search task, the mugshots, for which the vast majority of participants assumed a skimming behaviour. Additionally, as can be seen from Figure~\ref{fig:ac-skim}, our system achieved higher accuracy in search target prediction for participants who followed a specific search strategy than for those who skimmed most of the time. Well-performing participants also required fewer fixations to find the target (see Figure~\ref{fig:numfix}). Both findings are in line with previous works that describe eye movement control, i.e.\ the planning of where to fixate next, as an information maximisation problem~\cite{butko2010infomax,renninger2004information}. While participants unconsciously maximised the information gain by fixating appropriately during search, in some sense, they also maximised the information available for our learning method, resulting in 
higher 
prediction accuracy.

\section{Conclusion}

In this paper we demonstrated how to predict the search target during visual search from human fixations in an open-world setting.
This setting is fundamentally different from settings investigated in prior work, as we no longer assume that we have fixation data to train for these targets.
To address this challenge, we presented a new approach that is based on learning compatibilities between fixations and potential targets.
We showed that this formulation is effective for search target prediction from human fixations.
These findings open up several promising research directions and application areas, in particular gaze-supported image and media retrieval as well as human-computer interaction.
Adding visual behaviour features and temporal information to improve performance is a promising extension that we are planning to explore in future work. 

\section*{Acknowledgements}

This work was funded in part by the Cluster of Excellence on Multimodal Computing and Interaction (MMCI) at Saarland University.

{\small
\bibliographystyle{ieee}
\bibliography{references}
}
\end{document}